%% file: main.tex
\pdfoutput=1

\documentclass[11pt]{article}

\usepackage[preprint]{coling}

\usepackage{times}
\usepackage{latexsym}
\usepackage{multirow}
\usepackage{graphicx}
\usepackage{booktabs}
\usepackage{amsmath}
\usepackage{xspace}
\usepackage{color}

\usepackage{enumitem}
\setitemize[1]{itemsep=0pt,partopsep=0pt,parsep=\parskip,topsep=5pt,leftmargin=10pt}

\usepackage[T1]{fontenc}

\usepackage[utf8]{inputenc}

\usepackage{microtype}

\usepackage{inconsolata}

\usepackage{graphicx}

\usepackage{pifont}
\newcommand{\cmark}{\ding{51}}
\newcommand{\xmark}{\ding{55}}

\newcommand\eg{\textit{e.g.}}
\newcommand\ie{\textit{i.e.}}

\newcommand\wrt{\textit{w.r.t.}}

\title{Efficient Tool Use with Chain-of-Abstraction Reasoning}

\author{\textbf{Silin Gao$^{1,2\ast}$, Jane Dwivedi-Yu$^{2}$, Ping Yu$^{2}$, Xiaoqing Ellen Tan$^{2}$,} \\
\textbf{Ramakanth Pasunuru$^{2}$, Olga Golovneva$^{2}$, Koustuv Sinha$^{2}$} \\
\textbf{Asli Celikyilmaz$^{2}$, Antoine Bosselut$^{1\dagger}$, Tianlu Wang$^{2\dagger}$} \\
$^1$EPFL, $^2$FAIR @ Meta\\
{\tt $^1$\{silin.gao,antoine.bosselut\}@epfl.ch} \\
{\tt $^2$\{silingao,janeyu,pingyu,ellenxtan\}@meta.com} \\
{\tt $^2$\{rpasunuru,olggol,koustuvs,aslic,tianluwang\}@meta.com}
}

\begin{document}
\maketitle
\renewcommand{\thefootnote}{\fnsymbol{footnote}}
\footnotetext[1]{Work done during internship at FAIR.}
\footnotetext[2]{Equal Supervision.}
\renewcommand{\thefootnote}{\arabic{footnote}}
\begin{abstract}
To achieve faithful reasoning that aligns with human expectations, large language models (LLMs) need to ground their reasoning to real-world knowledge (\eg{}, web facts, math and physical rules).
Tools help LLMs access this external knowledge, but there remains challenges for fine-tuning LLM agents (\eg{}, Toolformer) to invoke tools in multi-step reasoning problems, where inter-connected tool calls require holistic and efficient tool usage planning.

In this work, we propose a new method for LLMs to better leverage tools in multi-step reasoning.
Our method, Chain-of-Abstraction (CoA), trains LLMs to first decode reasoning chains with abstract placeholders, and then call domain tools to reify each reasoning chain by filling in specific knowledge.
This planning with abstract chains enables LLMs to learn more general reasoning strategies, which are robust to shifts of domain knowledge (\eg{}, math results) relevant to different reasoning questions.
It also allows LLMs to perform decoding and calling of external tools in parallel, which avoids the inference delay caused by waiting for tool responses.
In mathematical reasoning and Wiki QA domains, we show that our method consistently outperforms previous chain-of-thought and tool-augmented baselines on both in-distribution and out-of-distribution test sets, with an average $\sim6\%$ absolute QA accuracy improvement.
LLM agents trained with our method also show more efficient tool use, with inference speed being on average $\sim$$1.4\times$ faster than baseline tool-augmented LLMs.
\end{abstract}

\input{sections/intro}

\input{sections/related_work}

\input{sections/method}

\input{sections/settings}

\input{sections/results}

\input{sections/conclusion}

\section*{Limitations}
We acknowledge a few limitations in our work.
First, datasets used for testing our method cannot have exhaustive coverage of all real-world reasoning scenarios.
We instead consider two representative reasoning domains, \ie{}, mathematical reasoning and general open-domain (Wikipedia) QA, and use English as a primary language in our testing.
Furthermore, our method is tested on the setting of fine-tuning the full LLMs, which requires considerable computational resources, while more efficient model training schemes, \eg{}, LoRA \citep{hu2021lora}, can be applied in future work.

\section*{Acknowledgements}
We thank Beatriz Borges, Gail Weiss, Syrielle Montariol, Li Mi and Zeming Chen for reading and providing comments on drafts of this paper.
Antoine Bosselut gratefully acknowledges the support of the Swiss National Science Foundation (No. 215390), Innosuisse (PFFS-21-29), the EPFL Science Seed Fund, the EPFL Center for Imaging, Sony Group Corporation, and the Allen Institute for AI.

\bibliography{main}

\input{sections/appendix}

\end{document}

%% file: sections/intro.tex
\begin{figure}[t]
\centering
\includegraphics[width=1.0\columnwidth]{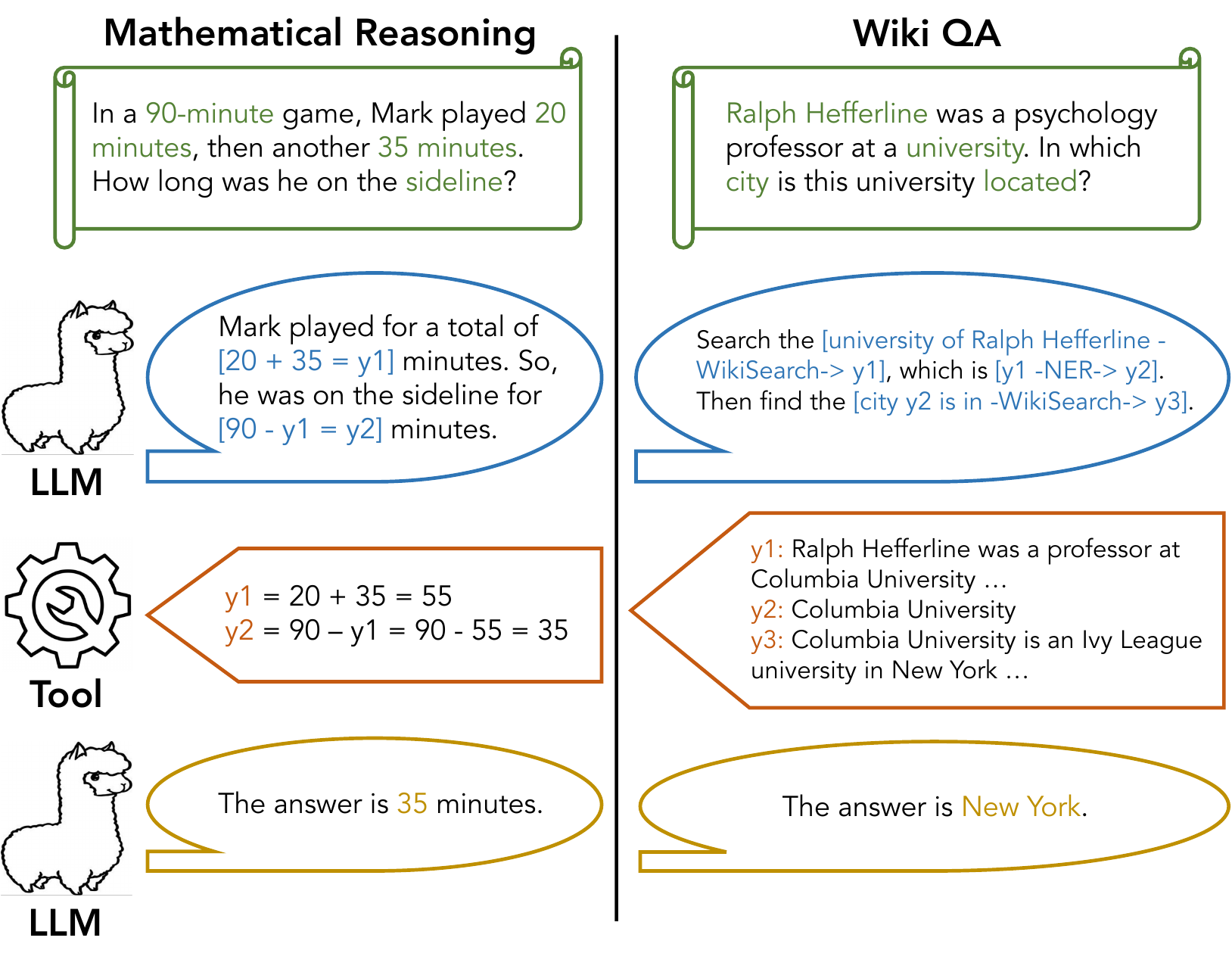}
\caption{Overview of chain-of-abstraction reasoning with tools. Given a domain question (green scroll), a LLM is fine-tuned to first generate an abstract multi-step reasoning chain (blue bubble), and then call external tools to reify the chain with domain-specific knowledge (orange label). The final answer (yellow bubble) is obtained based on the reified chain of reasoning.}
\label{method_overview}
\end{figure}

\section{Introduction}
Recent large language models (LLMs; \citealp{touvron2023llama2,anil2023palm2,openai2023gpt4}), have made progress at interpreting and executing instructions \citep{wei2021finetuned,chung2022scaling}, but still make errors when recalling and composing world knowledge for their responses, \eg{}, making unfactual statements \citep{maynez2020faithfulness,ji2023survey}, incorrect calculations \citep{patel2021nlp}, etc. 
Using auxiliary tools (\eg{}, a search engine to provide credible facts, a calculator for accurate math operations, etc.) at inference time can mitigate some of these errors, motivating tool-augmented language models that integrate external API calls into their output generations \citep{parisi2022talm,schick2023toolformer,hao2023toolkengpt}.

However, we show that current tool-augmented LLMs, \eg{}, Toolformer \citep{schick2023toolformer}, struggle to reliably and efficiently leverage tools in multi-step reasoning.
In particular, tool calls in multi-step reasoning tasks are often interleaved (\ie{}, the response of an API call is often part of the query of a subsequent call; as shown in Figure~\ref{method_overview}).
Without explicitly modeling these interconnections in reasoning chains, LLMs do not learn effective planning for tool use, which leads to less accurate reasoning with tools.\footnote{as verified by our analysis in \S\ref{results_analysis}}
Meanwhile, interleaving text generation with API calls also introduces inefficient inference ``waiting times,'' where the model must wait for the response from the API call before resuming the decoding process. This inefficiency becomes more significant in multi-step reasoning scenarios, when multiple rounds of API calls are typically required for each reasoning process.

In this work, we propose \textbf{C}hain-\textbf{o}f-\textbf{A}bstraction (\textbf{CoA}) reasoning, a robust and efficient method for LLMs to perform multi-step reasoning with tools.
As shown in Figure~\ref{method_overview}, LLMs are fine-tuned with a goal of making reasoning chains with abstract placeholders.
The placeholders do not affect LLMs' reasoning flow, and are subsequently infilled with specific knowledge retrieved from specialized tools, to ground the final answer generations.
Planning abstract chain of reasoning encourages LLMs to inter-connect multiple tool calls and adopt more feasible reasoning strategies, which are robust to the variation of domain knowledge involved in each reasoning process, \eg{}, specific calculation results.
Unlike previous methods where LLM decoding and API calls are executed in an interleaved manner, our method leverages tools to infill knowledge \textbf{once} after the whole chain of reasoning is generated.
This enables more efficient decoding across multiple examples (\eg, as in a stream) because CoA traces for subsequent examples can be decoded while tool calls are made for the preceding ones, amortizing overall inference time.
We develop a simple pipeline to build fine-tuning data for models to learn CoA, where we first prompt LLMs to re-write existing responses to instructions as abstract chains, and then use domain tools to check the validity of re-writing, as shown in Figure~\ref{method_rewrite}.

After training LLMs to learn CoA reasoning, we evaluate the finetuned models on two representative multi-step reasoning domains, including mathematical reasoning \citep{cobbe2021training,miao2020diverse,patel2021nlp,koncel2016mawps}, and Wikipedia (Wiki) QA \citep{yang2018hotpotqa,berant2013semantic,kwiatkowski2019natural,joshi2017triviaqa} that involves reasoning on factual descriptive knowledge.
We show that our method boosts LLMs' performances, with average $\sim$$7.5\%$ and $4.5\%$ absolute accuracy improvements on math and Wiki QA, respectively.
These improvements are consistent across both in-distribution
and (zero-shot) out-of-distribution test sets, and are especially pronounced on questions that require complex chain-of-thought reasoning.\footnote{\eg{}, more than 3 steps of math derivations} 
Meanwhile, our method also uses tools more efficiently than previous augmentation methods, with average $\sim$$1.47\times$ and $1.33\times$ faster inference speeds on math and Wiki QA tasks, respectively.
Finally, extensive human evaluation demonstrates that our method guides LLMs to learn more accurate reasoning, which leads to $\sim8\%$ fewer reasoning errors.

%% file: sections/related_work.tex
\section{Related Work}

\paragraph{Tool-Augmented LLMs}
There is growing interest in augmenting LLMs using external tools.
Considerable work has tried to adapt LLMs as tool-using reasoners through in-context learning, demonstrating promising performance improvements in various applications, \eg{}, math problem solving \citep{gao2023pal,chen2022program}, biomedical question answering \citep{jin2023genegpt} and self-critiquing \citep{gou2023critic}.
Nevertheless, guiding LLMs to effectively use tools using in-context demonstrations is challenging, which requires elaborate task-specific prompt engineering and is restricted by the model's instruction following ability \citep{jacovi2023comprehensive}. Noticing the limitations of in-context learning, several works teach LLMs to learn the usage of tools by fine-tuning \citep{parisi2022talm,schick2023toolformer,hao2023toolkengpt}, which more robustly improves LLMs' performance.
However, all above approaches adopt sequential interactions with tools throughout reasoning, slowing the inference speed as a function of the latency of the tool (or API) and the number of API calls that are made.

Some other prior works focus on using LLMs for multi-step reasoning with other modules.
In particular, ReAct \citep{yao2023react} and FireAct \citep{chen2023fireact} integrate LLMs with tools into a closed loop of thought, action and observation steps.
This verbose reasoning loop slows down the LLM decoding, and still incorporates tools via sequential interactions, resulting in inefficient inference.
Another line of work, Program of Thoughts \citep{chen2022program}, DECLARATIVE \citep{he2023solving} and PAL \citep{gao2023pal} prompt LLMs to generate program-based reasoning and interact with code executors, which however heavily rely on closed source coding models, \ie{}, Codex \citep{chen2021evaluating}, and are restricted to procedural arithmetic reasoning.
Building on these works, CoA proposes a framework to convert natural language reasoning traces into abstract representations, and uses the abstract reasoning traces as fine-tuning data to improve tool-augmented LLMs.
CoA also accelerates tool-augmented reasoning, by holistically planning the CoA traces and calling tools only once at inference time.

\paragraph{Tool Usage Planning}
Several previous works research tool usage planning in LLMs.
Specifically, HuggingGPT \citep{shen2023hugginggpt}, Chameleon \citep{lu2023chameleon}, OpenAGI \citep{ge2023openagi} and MetaTool \citep{huang2023metatool} focus on planning the high-level sequence of using multiple tools to address multi-domain mixed tasks.
Similarly, LATM \citep{cai2023large}, ML-BENCH \citep{liu2023ml} and Gorilla \citep{patil2023gorilla} aim at planning program-level integration of multiple APIs for designing scripts of procedural tasks, \eg{}, a script for training a model described by a GitHub repository.
ToolChain* \citep{zhuang2023toolchain} combines the planning of tool usage with tree-search-based reasoning \citep{yao2023tree,hao2023reasoning}, which is especially useful for procedural tasks \citep{xu2023tool,cobbe2021training}.
Different from above work, we focus on the planning of general chain-of-thought \citep{wei2022chain} reasoning with awareness of domain specialized tools.

%% file: sections/method.tex
\section{Method}


\begin{figure}[t]
\centering
\includegraphics[width=1.0\columnwidth]{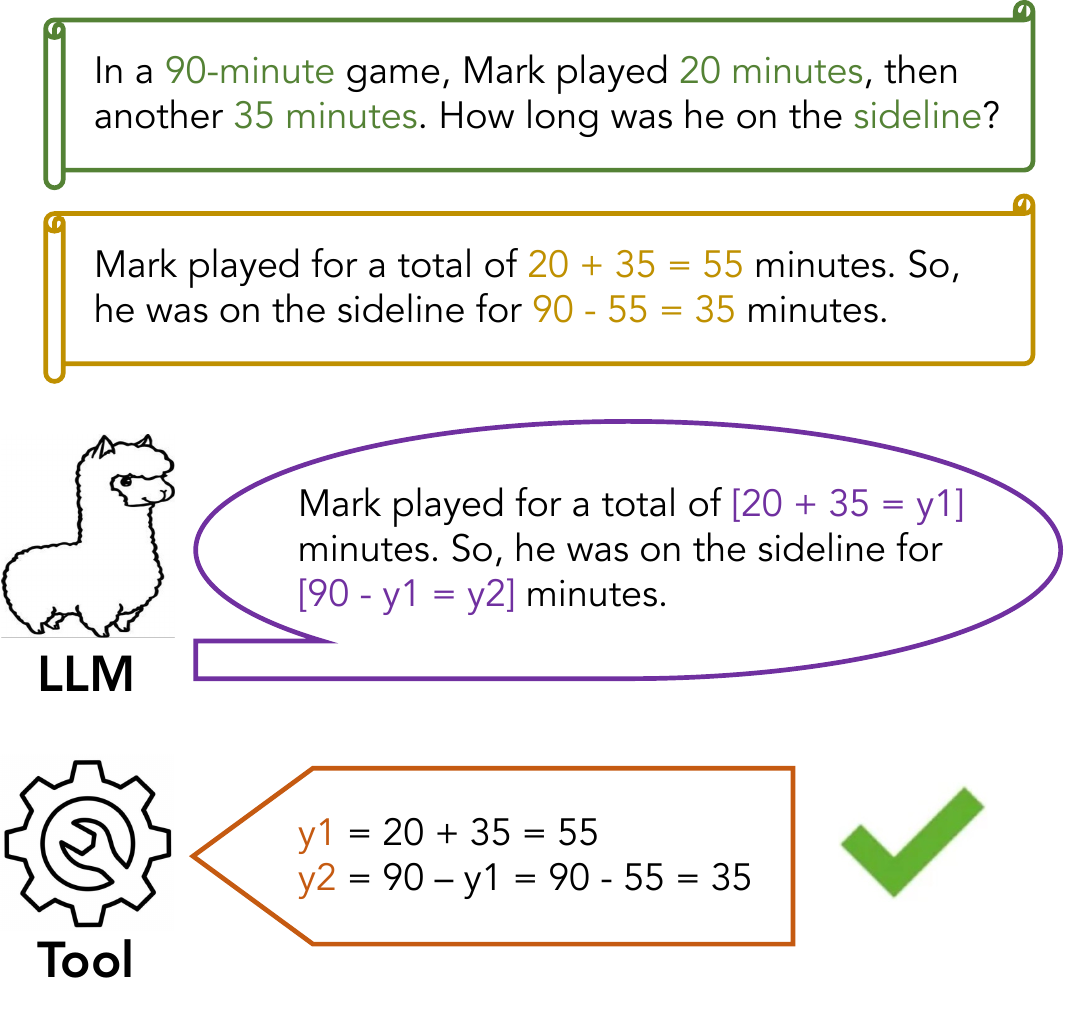}
\caption{Illustration of gold data re-writing for fine-tuning data construction. Given a pair of domain question (green scroll) and gold answer (yellow scroll), an LLM is prompted to re-write the gold answer as a reasoning chain with abstract variables (purple bubble). Then, domain specialized tools validate the correctness of the re-writing by checking whether the abstract chain can be reified to get the final answer (orange label).
}
\label{method_rewrite}
\end{figure}

\paragraph{Chain-of-Abstraction (CoA) Reasoning}

Our method decouples the general reasoning of LLMs from domain-specific knowledge obtained from external tools.
Figure~\ref{method_overview} shows an overview of our method.
In particular, we first fine-tune LLMs to generate reasoning chains with abstract placeholders, \eg{}, $y1$, $y2$ and $y3$,\footnote{We also test placeholders in single-character format, \eg{}, $x$, $y$ and $z$, but these led to sub-optimal results.} as shown in Figure~\ref{method_overview}.
In the second stage, we reify each reasoning chain by replacing placeholders with domain-specific knowledge obtained from external tools, \eg{}, calculation results from a calculator, relevant articles retrieved from web search engine, etc.
Finally, the question is answered based on the reified reasoning chain.

Note that since the LLMs are trained to generate abstract chain of reasoning instead of regular chain-of-thought (CoT) reasoning with explicit values, this enables LLMs to focus on learning general and holistic reasoning strategies without needing to generate instance-specific knowledge for the model's parameters.  
Moreover, decoupling general reasoning and domain-specific knowledge enables LLM decoding to proceed and switch between different samples in parallel with API calling (via a pipeline), \ie{}, LLM can start generating the next abstract chain while the tool fills the current chain, which speeds up the overall inference process.

\paragraph{Fine-tuning Data Construction}

To construct chain-of-abstraction (CoA) data for fine-tuning LLMs, we collect question answering (QA) samples from existing open-source QA datasets \citep{cobbe2021training,miao2020diverse,yang2018hotpotqa}, and prompt LLaMa-70B \citep{touvron2023llama} to re-write the answer of each sampled question, as shown in Figure~\ref{method_rewrite}.
Specifically, we prompt LLaMa-70B to label the spans in gold answers that correspond to knowledge operations (\eg{}, math derivations, statements based on Wikipedia references) and then to re-write the sentences with labeled spans as fillable CoA traces, where the operation results are replaced with abstract placeholders.
For example, the two derivations in the example in Figure~\ref{method_rewrite} are re-written as ``[$20 + 35 = y1$]" and ``[$90 - y1 = y2$]", respectively.

Note that an intermediate knowledge operation result may appear multiple times in an answer, \eg{}, in Figure~\ref{method_rewrite}, the first equation's result $55$ is used in the second equation.
We prompt LLaMa-70B to replace all occurrences of the same intermediate result with the same placeholder, thereby explicitly connecting the multiple reasoning steps.
To ensure that the re-written data is accurate, we use domain-specialized tools to verify the correctness of each CoA reasoning trace.\footnote{Detailed implementations of reasoning chain verification are described in Sec.~\ref{ss:math} and \ref{ss:wikiqa}.}
Specifically, we use the tools to execute the labeled operations in each CoA, and only keep questions whose CoA can be infilled with valid results by the tools.


%% file: sections/settings.tex
\section{Experimental Settings}
We conduct our experiments on two representative domains: mathematical reasoning and Wikipedia (Wiki) QA, which involves commonsense and logical reasoning on factual descriptive knowledge.

\subsection{Mathematical Reasoning}
\label{ss:math}
Given a math question, the QA system needs to generate a natural language solution to the problem with step-by-step arithmetic derivations (as demonstrated in the left column of Figure~\ref{method_overview}).
We assume that the derivations involved in the solution are the specialized knowledge operations required in this domain, which are labeled in square brackets with derivation results being replaced by abstract placeholders, e.g., ``[$20 + 35 = y1$]".

\paragraph{Datasets}
We construct most of our fine-tuning CoA data by re-writing the GSM8K \citep{cobbe2021training} training set, which contains 7473 linguistically diverse grade school math problems.
As GSM8K dataset focuses on multi-step reasoning, it lacks coverage of single-step arithmetic problems, so we also re-write an additional set of 691 single-step math problems from the ASDiv \citep{miao2020diverse} dataset.
Across these re-written datasets, we find that $\sim76.6\%$ of the CoA reasoning traces generated by LLaMa-70B are verified by our equation solver (described below). 
Table~\ref{tab:ft-data-math} shows the reasoning step distribution (\ie{}, number of derivations) of our constructed fine-tuning data.

\begin{table}[t]
\centering
\resizebox{1.0\columnwidth}{!}{
\smallskip\begin{tabular}{lccccccc}
\toprule
\multirow{2}*{\textbf{Source}}  &  \multicolumn{7}{c}{\textbf{Reasoning Step}}   \\
                                        \cmidrule(lr){2-8}
                                        &  1  &   2  &   3  &  4  &  5  &  $>$5 & All \\
\toprule
GSM8K               &  8  & 1540 & 1648 & 1164 & 666  & 553 & 5579 \\
ASDiv               & 677 &   0  &   0  &   0  &   0  &  0  & 677  \\
\bottomrule
\end{tabular}
}
\caption{Reasoning step distribution of correctly re-written reasoning chains in math domain.}
\label{tab:ft-data-math}
\end{table}


For an in-distribution evaluation, we test models on GSM8K and ASDiv, containing 1319 and 2305 testing problems.
To further test the models' generalization ability, we also conduct zero-shot evaluation on other representative math datasets, including SVAMP \citep{patel2021nlp} and MAWPS \citep{koncel2016mawps}, which contain 1000 and 2065 testing samples, respectively.\footnote{For the MAWPS benchmark, we test on the 395, 508, 562 and 600 math problems from AddSub, SingleEq, SingleOp and MultiArith portions, respectively.}

\paragraph{Domain Tool}
We use an equation solver to perform the arithmetic derivations required in the math domain.
Our equation solver first extracts the derivations labeled in the CoA reasoning, e.g., ``[$20 + 35 = y1$]" and ``[$90 - y1 = y2$]", and combines all derivations into a system of equations.
Then the system of equations is solved by the SymPy toolkit,\footnote{\url{https://www.sympy.org/en/index.html}} to get the true value of each variable (\ie{}, the value of the abstract placeholder).
Finally, our equation solver returns the reified chain of reasoning by replacing all the variables with their solved true values (including the final answer).

\begin{table}[t]
\centering
\resizebox{1.0\columnwidth}{!}{
\smallskip\begin{tabular}{ll}
\toprule
\multirow{2}*{\textbf{Question}} & The director of the romantic comedy ``Big Stone Gap'' is based in \\
& what New York city? \\
\midrule
\textbf{Answer} & Greenwich Village \\
\midrule
\multirow{3}*{\textbf{Wikipedia}} & Big Stone Gap (film) > Big Stone Gap is a 2014 American romantic \\
\multirow{3}*{\textbf{References}} & comedy film directed by Adriana Trigiani. \\
 & Adriana Trigiani > Adriana Trigiani is an Italian American film \\
 & director based in Greenwich Village. \\
\midrule
\multirow{3}*{\textbf{CoA Trace}} & Find the [director of romantic comedy ``Big Stone Gap'' -Wiki-> y1]. \\
& The name of this film's director is [y1 -NER(person)-> y2]. \\
& Then determine [y2 in what New York city -Wiki-> y3]. \\
\bottomrule
\end{tabular}
}
\caption{Example of CoA fine-tuning data construction in Wiki QA domain.}
\label{tab:rewrite_example_wiki}
\end{table}

\subsection{Wikipedia QA}
\label{ss:wikiqa}
Given a question based on Wikipedia knowledge, the model needs to first identify Wikipedia articles as references related to the question, and then reason on key knowledge in the reference articles to answer the question (as shown in the right column of Figure~\ref{method_overview}).
We assume that the specialized knowledge operation in this domain is the retrieval of relevant Wikipedia articles and important named-entities, which are re-written as Wikipedia searching (WikiSearch) and named-entity recognition (NER)\footnote{We use NER to extract entities from the article that bridge the former WikiSearch results to the latter WikiSearch queries.} queries.
Table~\ref{tab:rewrite_example_wiki} shows an example of a re-written CoA trace for Wiki QA.

\paragraph{Datasets}
We use the HotpotQA \citep{yang2018hotpotqa} dataset to construct our fine-tuning CoA data in the Wiki QA domain.
HotpotQA contains 113K multi-hop QA examples, each labeled with two Wikipedia articles that provide supporting knowledge.
Among the 90447 training QA pairs, we identify 72991 as \textbf{Bridge} QA pairs, where an intermediate entity must be identified to link the answer to the question, as shown in Table~\ref{tab:rewrite_example_wiki}.
The remaining 17456 are \textbf{Comparison} QA pairs, where the attributes of two entities are compared, \eg{}, ``Are Randal Kleiser and Kyle Schickner of the same nationality?''.
We prompt LLaMa-70B to re-write these training QAs into CoAs with WikiSearch and NER queries, and verify each CoA with our domain tools (described below), by checking whether all the articles returned by the WikiSearch queries match one of the titles in the gold articles.
Finally, 8956 Bridge QAs and 5405 Comparison QAs are used as fine-tuning data, whose re-written CoAs pass the verification.\footnote{Compared to mathematical reasoning, generating CoA data for Wiki QA requires more complex tool use that combines WikiSearch and NER models, leading to a lower re-writing success rate ($\sim15.9\%$).}
For Wiki QA, we note that besides training a LLM to produce CoA data using WikiSearch, we also fine-tune a second LLM to learn to generate the final gold answer based on a correctly reified CoA reasoning trace.

We evaluate models on the HotpotQA development set, which contains 5918 Bridge QA pairs and 1487 Comparison QA pairs. Similar to the mathematical reasoning domain, we also conduct zero-shot evaluation on other open-domain QA datasets: WebQuestions (WQ; \citealp{berant2013semantic}), NaturalQuestions (NQ; \citealp{kwiatkowski2019natural}) and TriviaQA \citep{joshi2017triviaqa}, which contain 2032, 3610 and 17944 test questions, respectively.

\paragraph{Domain Tools}
The specialized tools required for Wiki QA include a Wikipedia search engine to retrieve reference articles, and a NER toolkit to extract entities that bridge multi-step searching queries.
We follow Toolformer \citep{schick2023toolformer} and implement a Wikipedia search engine as a BM25 retriever \citep{robertson1995okapi,baeza1999modern} that indexes the Wikipedia dump from the KILT benchmark \citep{petroni2021kilt}.
We use the BM25 retriever to search the top-10 articles relevant to the input query, and then re-rank the articles based on their Sentence-BERT \citep{reimers2019sentence} embedding cosine similarity with the question.
After re-ranking, the top-$1$ article is selected to be the final search result.

We use SpaCy\footnote{\url{https://spacy.io/models/en}} (\texttt{en\_core\_web\_sm}) as the NER toolkit to extract named entities.
To simplify NER, we aggregate the numerous SpaCy NER types into 6 general classes, as shown in Table~\ref{tab:ner_aggr}.
If multiple named entities are recognized, we input each recognized entity to the subsequent WikiSearch query, and select the entity whose subsequent search result has the highest Sentence-BERT embedding cosine similarity with the question.

\begin{table}[t]
\centering
\resizebox{1.0\columnwidth}{!}{
\smallskip\begin{tabular}{cl}
\toprule
\textbf{General}  &  \multirow{2}*{\textbf{SpaCy NER Types included in each General Class}}   \\
\textbf{Class}  &  \\
\toprule
person  & PERSON  \\
group  & NORP, ORG, LANGUAGE \\
location  & GPE, FAC, LOC \\
culture & EVENT, WORK\_OF\_ART, LAW, PRODUCT \\
date  & DATE, TIME \\
numeral & CARDINAL, PERCENT, MONEY, QUANTITY, ORDINAL \\
\bottomrule
\end{tabular}
}
\caption{Aggregation of SpaCy NER types.}
\label{tab:ner_aggr}
\end{table}

\begin{table*}[t]
\centering
\resizebox{1.0\textwidth}{!}{
\smallskip\begin{tabular}{ll@{~}ccccccccc}
\toprule
\multirow{2}*{\textbf{Model}} & \multirow{2}*{\textbf{Method}} & \multirow{1.8}*{\textbf{Use}} & \multirow{2}*{\textbf{GSM8K}} & \multirow{2}*{\textbf{ASDiv}} & \multirow{2}*{\textbf{SVAMP}} & \multicolumn{5}{c}{\textbf{MAWPS}} \\
            \cmidrule(lr){7-11}
 &  & \textbf{Tool} &  &  &  & \textbf{AddSub} & \textbf{SingleEQ} & \textbf{SingleOp} & \textbf{MultiArith} & \textbf{All} \\
\toprule
\multirow{4}*{\textbf{LLaMa-2}} & CoT-FSP & \multirow{2}*{\xmark} & 16.38 & 47.85 & 38.40 & 52.41 & 63.39 & 82.03 & 43.33 & 60.53 \\
\multirow{4}*{\textbf{-7B}} & CoT-FT &  & \underline{35.33} & \underline{57.18} & \underline{48.20} & \underline{66.08} & \underline{74.41} & \underline{85.23} & \underline{65.00} & \underline{73.03}  \\
            \cmidrule(lr){2-11}
            & Toolformer & \multirow{2}*{\cmark} & 17.59 & 48.55 & 37.10 & 47.34 & 58.46 & 79.54 & 50.67 & 59.81 \\
            & CoA  &  & \textbf{37.83}$^{\ast}$ & \textbf{57.61} & \textbf{51.70}$^{\ast}$ & \textbf{72.15}$^{\ast}$ & \textbf{82.48}$^{\ast}$ & \textbf{86.48}$^{\ast}$ & \textbf{73.17}$^{\ast}$ & \textbf{78.89}$^{\ast}$ \\
\midrule
\multirow{5}*{\textbf{LLaMa-2}} & CoT-FSP  & \multirow{3}*{\xmark} & 24.03 & 54.14 & 51.30 & \underline{71.90} & 72.44 & 85.41 & 74.00 & 76.32 \\
\multirow{5}*{\textbf{-Chat-7B}}  & CoT-FT &  & 35.41 & 59.00 & 46.90 & 58.23 & 72.24 & 85.41 & 73.00 & 73.37  \\
  & CoA (no Tool) &  & 35.03 & 58.79 & \underline{51.50} & 68.10 & \underline{74.21} & \underline{86.48} & \underline{77.67} & \underline{77.38} \\
  \cmidrule(lr){2-11}
  & Toolformer  & \multirow{3}*{\cmark} & 23.65 & 50.85 & 48.80 & 61.01 & 69.09 & 81.85 & 68.50 & 70.85  \\
  & Toolformer - Math &  & \underline{36.01} & \underline{59.18} & 47.60 & 58.99 & 72.44 & 85.94 & 75.50 & 74.43 \\
 & CoA  &  & \textbf{38.29}$^{\ast}$ & \textbf{59.57} & \textbf{54.20}$^{\ast}$ & \textbf{72.41} & \textbf{81.89}$^{\ast}$ & \textbf{88.26}$^{\ast}$ & \textbf{83.00}$^{\ast}$ & \textbf{82.13}$^{\ast}$ \\
\midrule
\multirow{4}*{\textbf{LLaMa-2}} & CoT-FSP  & \multirow{2}*{\xmark} & 56.18 & 65.94 & 70.60 & 86.08 & 89.17 & \underline{92.88} & 84.50 & 88.23 \\
\multirow{4}*{\textbf{-Chat-70B}} & CoT-FT  &  & 60.50 & 70.24 & 70.40 & 81.52 & 87.60 & 92.35 & 89.17 & 88.18  \\
  \cmidrule(lr){2-11}
                                 & Toolformer & \multirow{3}*{\cmark} & 52.54 & 69.07 & \textbf{73.60} & \textbf{86.84} & 89.76 & 91.46 & 81.50 & 87.26 \\
             & Toolformer - Math &  & \underline{61.03} & \underline{70.59} & 73.20 & 85.57 & \underline{91.34} & 91.99 & \underline{92.00} & \underline{90.60} \\
            & CoA  &  & \textbf{62.32}$^{\ast}$ & \textbf{71.89}$^{\ast}$ & \underline{73.40} & \underline{86.33} & \textbf{94.49}$^{\ast}$ & \textbf{93.06} & \textbf{92.33} & \textbf{91.91}$^{\ast}$ \\
\bottomrule
\end{tabular}
}
\caption{Evaluation results on LLaMa-2 and LLaMa-2-Chat for mathematical reasoning. ``All'' denotes the averaged results on four MAWPS portions. Exact match rate to the final gold answer (\ie{}, accuracy) is reported. 
For each base model, the best and second-best results are \textbf{bolded} and \underline{underlined}, respectively. The best results labeled with $^{\ast}$ are significantly better than their corresponding second-best results, with the significant test p-value $< 0.05$.
}
\label{tab:math_results}
\end{table*}

\subsection{Baselines}

We apply our \textbf{CoA} reasoning method to both 7B and 70B LLaMa models, and test various model versions including the first version of LLaMa \citep{touvron2023llama} and the more advanced LLaMa-2 and LLaMa-2-Chat \citep{touvron2023llama2}.
We compare our method to several baselines, including: a) few-shot prompting using 8 randomly sampled QA exemplars from the original (\ie{}, not re-written) chain-of-thought data (\textbf{CoT-FSP}), b) fine-tuning with original chain-of-thought data (\textbf{CoT-FT})\footnote{Note that in Wiki QA domain, the HotpotQA data used for prompting or fine-tuning baselines is pre-processed to contain both gold Wikipedia articles (serving as chain-of-thought explanations) and the final answer.}, and c) \textbf{Toolformer} \cite{schick2023toolformer} which fine-tunes LLMs on CCNet \citep{wenzek2020ccnet} texts augmented with API calls.
For evaluation on Wiki QA, we also compared our method with \textbf{FireAct} \citep{chen2023fireact}, which fine-tunes LLMs on HotpotQA ReAct \citep{yao2023react} trajectories distilled from GPT-4 \citep{openai2023gpt4}.

%% file: sections/results.tex
\section{Results and Analysis}
\label{results_analysis}
\subsection{Mathematical Reasoning}
\label{math_qa_exp}
Table~\ref{tab:math_results} shows the evaluation results for the LLaMa-2 and LLaMa-2-Chat models.\footnote{We include similar evaluation results for the original LLaMa model (7B) in Appendix \ref{apdx:full_results}.}
On the GSM8K and ASDiv datasets, our CoA method outperforms the few-shot baseline CoT-FSP and the regular fine-tuning baseline CoT-FT, demonstrating that CoA fine-tuning with tool augmentation is more effective in adapting LLMs to multi-step reasoning tasks.
Similarly, when evaluated on out-of-distribution datasets, SVAMP and MAWPS, CoA also consistently outperforms the baselines.
Interestingly, for these out-of-distribution datasets, CoT-FT lags further behind CoA, particularly for 7B models, showing that CoA reasoning yields more distributionally robust reasoning performance.

Our CoA method also surpasses the tool-augmented baseline Toolformer, which implies that planning the abstract variables in CoA can improve the accuracy of reasoning with tools.
However, as Toolformer is not originally trained with in-domain fine-tuning data,\footnote{Toolformer is fine-tuned on CCNet data, which may not contain rich mathematical reasoning samples.} we also fine-tune a new version of Toolformer with the chain-of-thought data from GSM8K and ASDiv, denoted as \textbf{Toolformer - Math} in Table~\ref{tab:math_results}.
We also observe that CoA performs better than Toolformer - Math, confirming that the introduction of abstract variables enables more robust tool use compared to direct integration of API calls within chain-of-thought reasoning.

\paragraph{Ablation Study}
We verify that the robust generalization performance of our CoA method does not merely benefit from using additional tools, by fine-tuning another LLM to solve the equation (from the same model backbone), rather than calling the equation solver, denoted as \textbf{CoA (no Tool)} in Table~\ref{tab:math_results}. 
We find that CoA (no Tool) performs consistently worse than CoA across all datasets, confirming that using specialized tools enables LLM agents to conduct more precise operations, rather than directly solving the same operations.
However, CoA (no Tool) still outperforms all baseline methods on zero-shot generalization to SVAMP and MAWPS datasets, implying that learning abstract reasoning chains also contributes to better robustness of CoA, perhaps due to better planning of multiple reasoning steps indexed by abstract variables.

\begin{figure}[t]
\centering
\includegraphics[width=1.0\columnwidth]{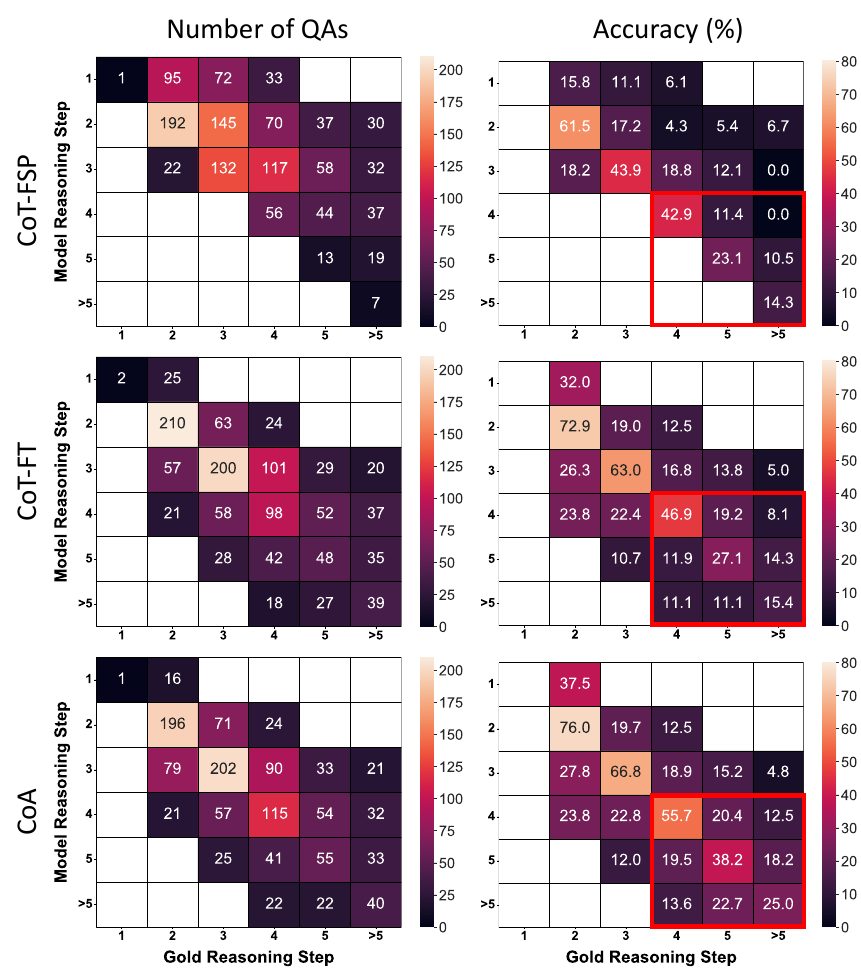}
\caption{GSM8K evaluation results on LLaMa-2-Chat-7B \wrt{} the number of reasoning steps in the predicted and gold reasoning chain. (Left) The number of test examples that belong to each stratum. (Right) The corresponding model accuracy (\%) for those examples. Non-diagonal cells with fewer than 15 examples are ignored.}
\label{math_heatmap}
\end{figure}

\paragraph{Reasoning Steps}
Our findings suggest that the benefits of chain-of-abstraction reasoning are most pronounced when problems require long reasoning chains to be solved. Figure~\ref{math_heatmap} shows the stratified performance of three models on GSM8K QA, relative to the number of reasoning steps in the predicted and gold reasoning chains. 
Compared to the few-shot CoT-FSP, CoA produces reasoning chains that more often match the length of the gold reasoning chains, as reflected by the heat-map statistics (left column) being more aggregated around the diagonal (comparable to CoT-FT).
At the same time, we observe that models achieve better QA accuracy when the number of reasoning steps in their generated answers are aligned with the gold references (\ie{}, the diagonal of heat-maps in right column).
Above results show that fine-tuned models are better at learning to produce reasoning chains that match the true reasoning chain for the problem.

\begin{table}[t]
\centering
\resizebox{0.75\columnwidth}{!}{
\begin{tabular}{lcc}
\toprule
\multirow{2}*{\textbf{Method}}  &  \multicolumn{2}{c}{\textbf{Error Rate}}   \\
                                \cmidrule(lr){2-3}
                                &  Arithmetic  &   Reasoning   \\
\toprule
CoT-FSP         & 17.3 & 70.3   \\
CoT-FT          & 25.2 & 67.8   \\
\midrule
CoA & \textbf{0.0} & \textbf{60.4} \\ 
\bottomrule
\end{tabular}
}
\caption{Human evaluation results of arithmetic and reasoning error rates on 200 GSM8K test samples. Models developed based on LLaMa-2-Chat-7B are presented.}
\label{tab:human_eval}
\end{table}

\begin{figure}[t]
\centering
\includegraphics[width=1.0\columnwidth]{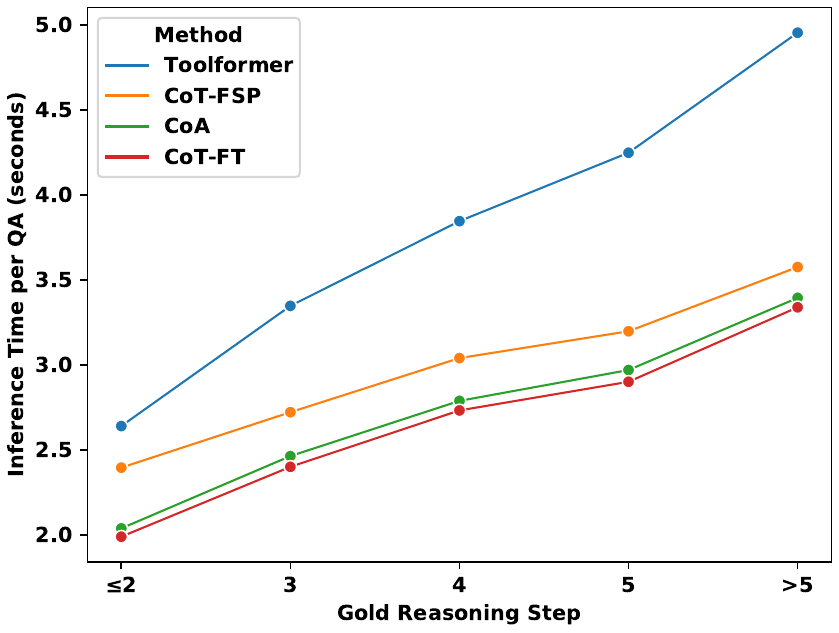}
\caption{Wall-clock inference time on GSM8K (seeded with LLaMa-2-Chat-7B). Average time of answering a question is measured (in seconds) \wrt{} the number of gold reasoning steps required for the question.}
\label{efficiency_lines}
\end{figure}

\begin{table}[t]
\centering
\begin{tabular}{lc}
\toprule
\textbf{Method}  &  \textbf{Accuracy}   \\
\toprule
CoT-FSP       & 27.90 \\
CoT-FT        & 39.12 \\
Toolformer    & 24.56 \\
Toolformer - Math & 35.25 \\
CoA & \textbf{40.79} \\
\bottomrule
\end{tabular}
\caption{
Evaluation results on GSM8K with self-consistency decoding (seeded with LLaMa-2-Chat-7B). Each model uses majority voting to aggregate the answers of 16 sampled reasoning chains
}
\label{tab:self_consistency}
\end{table}

\begin{table*}[t]
\centering
\resizebox{1.0\textwidth}{!}{
\smallskip\begin{tabular}{llcccccccc}
\toprule
\multirow{2}*{\textbf{Model}} & \multirow{2}*{\textbf{Method}} & \multirow{1.8}*{\textbf{Use}}  & \multicolumn{4}{c}{\textbf{HotpotQA}} & \multirow{2}*{\textbf{WQ}} & \multirow{2}*{\textbf{NQ}} & \multirow{2}*{\textbf{TriviaQA}} \\
            \cmidrule(lr){4-7}
 & & \textbf{Tool} & \textbf{Bridge} & \textbf{Comparison} & \textbf{Both} & \textbf{Time} &  &  &   \\
\toprule
\multirow{6}*{\textbf{LLaMa-2}}  & CoT-FSP  & \multirow{2}*{\xmark} & 11.69 & 45.46 & 18.47 & 2.074 & 34.65 & 30.91 & 53.48 \\
\multirow{6}*{\textbf{-Chat-7B}} & CoT-FT   &  & 14.24 & \underline{56.69} & 22.77 & \underline{1.937} & 33.51 & 25.40 & 51.05 \\
    \cmidrule(lr){2-10}
                                  & Toolformer  & \multirow{4}*{\cmark} & 12.99 & 44.59 & 20.00 & 2.350 & \underline{36.22} & 30.22 & 54.15 \\
                                  & Toolformer - Wiki &  & 15.68 & 56.42 & 23.86 & 2.301 & \textbf{36.61} & 32.96 & \underline{55.08} \\
                                  & FireAct   &  & \underline{19.18} & 54.14 & \underline{26.20} & 2.706 & 36.02 & \underline{35.87} & 52.96 \\
                                & CoA  &  & \textbf{21.00}$^{\ast}$ & \textbf{56.96} & \textbf{28.22}$^{\ast}$ & \textbf{1.896} & 35.97 & \textbf{38.67}$^{\ast}$ & \textbf{57.90}$^{\ast}$ \\
\midrule
\multirow{4}*{\textbf{LLaMa-2}}  & CoT-FSP   & \multirow{2}*{\xmark} & 21.39 & 56.62 & 28.47 & 6.668 & 34.89 & 37.42 & 63.61 \\
\multirow{4}*{\textbf{-Chat-70B}} & CoT-FT   &  & 23.84 & \underline{63.95} & 31.90 & \underline{6.401} & 34.15 & 39.75 & 62.28 \\
    \cmidrule(lr){2-10}
                                & Toolformer & \multirow{3}*{\cmark} & 22.24 & 56.09 & 29.04 & 6.888 & \underline{37.16} & 40.42 & 64.31 \\
                                & Toolformer - Wiki &  & \underline{26.38} & 63.82 & \underline{33.90} & 6.855 & \textbf{37.70} & \underline{41.25} & \underline{66.64} \\
                                & CoA &  & \textbf{27.61}$^{\ast}$ & \textbf{64.09} & \textbf{34.94}$^{\ast}$ & \textbf{6.369} & 36.37 & \textbf{43.57}$^{\ast}$ & \textbf{69.08}$^{\ast}$ \\
\bottomrule
\end{tabular}
}
\caption{Wiki QA evaluation results on LLaMa-2-Chat-based models. ``Both'' denotes the overall evaluation results on both bridge and comparison portions of HotpotQA. ``Time'' denotes the average seconds that each agent needs to answer a question in HotpotQA. Exact match rate to the final gold answer (\ie{}, accuracy) is reported.
For each base model, the best and second-best results are \textbf{bolded} and \underline{underlined}, respectively. The best results labeled with $^{\ast}$ are significantly better than their corresponding second-best results, with the significant test p-value $< 0.05$.
}
\label{tab:wiki_results}
\end{table*}

Interestingly, we find that CoA, compared to CoT-FT, achieves higher performance especially on questions that require more reasoning steps.
In the right column of Figure~\ref{math_heatmap}, CoA's improvement over CoT-FT is more pronounced on questions with more than $3$ steps in the gold reasoning chain (highlighted with red squares). 
This indicates that the model trained with CoA has more robust long chain-of-thought reasoning capability, which is learned from planning with abstractions.

\paragraph{Human Evaluation}
To more comprehensively verify that CoA improves both knowledge operation (\ie{}, arithmetic by using tools) and reasoning accuracy, we conduct a human evaluation on different model answers to 200 randomly sampled GSM8K test questions. 
Specifically, given a GSM8K question and a model's answer to the question, we ask human workers to judge whether the answer contains any arithmetic errors (\eg{}, wrong calculations, invalid equations) or reasoning errors unrelated to math derivations (\eg{}, misunderstanding of the question, improper strategy for solving the question), and report how often the model makes these two kinds of errors.
In Table~\ref{tab:human_eval}, we find that CoA effectively reduces arithmetic errors to zero, due to the use of equation solver to perform accurate calculations.
More importantly, our method also makes fewer reasoning errors compared to the baselines, verifying that CoA fine-tuning guides the model to learn more accurate reasoning through the holistic planning of abstract reasoning chains.
By contrast, ordinary fine-tuning (\ie, CoT-FT) produces a more limited reasoning improvement compared to the few-shot CoT-FSP, while also failing to suppress arithmetic errors.


\paragraph{Inference Efficiency}
Importantly, we find that the performance benefits of CoA reasoning do not come with increased computational costs.
In Figure~\ref{efficiency_lines}, we show the average time (seconds) that CoA and baseline agents (seeded with LLaMa-2-Chat-7B) needs to answer a question \wrt{} required gold reasoning steps.
Compared to the CoT baselines, CoA requires less time than the few-shot baseline CoT-FSP, whose generation needs to be conditioned on additional examples.
However, CoA is slightly less inference-efficient compared to CoT-FT, likely due to the decoding of additional tokens (\eg{}, ``['' and ``]'') for the abstract statements.

Compared to Toolformer, CoA has a lower and flatter inference time curve, indicating better scaling as the number of reasoning steps increases.
This difference arises because CoA decouples the generation of (abstract) reasoning chains from the retrieval of knowledge (\ie{}, tool use), allowing full reasoning chains to be decoded before any tool is called.
This procedure amortizes inference costs in two ways. 
First, tool calls are made after the CoA trace has been decoded, enabling parallel tool calls for the same trace (\eg, using an equation solver once rather than multiple calls to a calculator), and avoiding the time delay caused by waiting for external API responses. Consequently, the model fine-tuned with CoA is more efficient at multi-step reasoning, especially when the number of reasoning steps (\ie{}, tool calls) increases.
Second, across multiple examples, the model can generate the CoA trace of the next example while tool calls are made for the preceding one, parallelizing CoA decoding and tools calls across examples.

\paragraph{Self-Consistency Decoding}
Besides of greedy decoding, we also test more advanced inference strategy, \ie{}, self-consistency \citep{wang2022self} decoding, on our CoA reasoning method.
We test all methods on the GSM8K dataset seeded with LLaMa-2-Chat-7B.
Each method samples 16 reasoning chains and uses majority voting to aggregate the 16 answers derived by the reasoning chains, to get the final answer.
For the hyperparameters of sampling, we set the temperature, top-k and top-p as 1.0, 40 and 0.5, respectively.
Table~\ref{tab:self_consistency} shows our evaluation results.
We find that our CoA method consistently outperforms all baseline methods when shifting from greedy decoding to self-consistency decoding.
This shows that our method also has better potential to be generalized to different LLM decoding schemes.

\subsection{Wiki QA}
\label{wiki_qa_exp}
Table~\ref{tab:wiki_results} shows our Wiki QA results using LLaMa-2-Chat models.\footnote{We include similar evaluation results on LLaMa-2-7B in Appendix \ref{apdx:full_results}.}
Similar to mathematical reasoning, we fine-tune a new version of Toolformer with in-domain chain-of-thought data from HotpotQA, denoted as \textbf{Toolformer - Wiki}.
On HotpotQA, CoA achieves higher exact match rates with the gold reference compared to the few-shot or fine-tuning baselines.
In particular, CoA outperforms all baselines on the more challenging bridge-type QAs, where two steps of reasoning over Wikipedia knowledge are \textit{consecutively} entangled, \ie{}, cannot be performed independently in parallel as in comparison-type QAs. 
Compared to FireAct fine-tuning, CoA also achieves better performance on both bridge and comparison QAs, without requiring data distilled from closed source GPT-4.

As with mathematical reasoning, CoA agents also perform more efficient inference than Toolformer and FireAct agents when answering HotpotQA questions.
We also find that CoA is more efficient (\textbf{Time} column) than both CoT-FSP and CoT-FT, as CoA does not require few-shot examples as additional inputs and does not need to generate long Wiki articles, which are instead provided by the search engine.
Finally, CoA improves over the baseline methods in zero-shot generalization experiments on other Wiki QA datasets, outperforming all baselines on NaturalQuestions and TriviaQA, and matching the best baselines on WebQuestions.

%% file: sections/conclusion.tex
\section{Conclusion}
In this work, we propose to decouple the general reasoning of LLM agents from specialized knowledge obtained via external tools.
Our method, chain-of-abstraction (CoA), encourages LLMs to learn the planning of abstract multi-step reasoning, which are more robust to out-of-distribution knowledge shifts.
CoA also achieves a more efficient pipeline for tool usage that significantly improves the speed of tool-augmented multi-step reasoning.
The simple, yet effective, implementations of our method on two diverse tasks (\ie{}, math reasoning and open-domain QA) demonstrate its potential for being adapted to new reasoning scenarios.

%% file: sections/appendix.tex
\appendix


\begin{table*}[t]
\centering
\resizebox{1.0\textwidth}{!}{
\smallskip\begin{tabular}{llcccccccc}
\toprule
\multirow{2}*{\textbf{Model}} & \multirow{2}*{\textbf{Method}}  &  \multirow{2}*{\textbf{GSM8K}} & \multirow{2}*{\textbf{ASDiv}} & \multirow{2}*{\textbf{SVAMP}} & \multicolumn{5}{c}{\textbf{MAWPS}} \\
            \cmidrule(lr){6-10}
&  &  &  &  & \textbf{AddSub} & \textbf{SingleEQ} & \textbf{SingleOp} & \textbf{MultiArith} & \textbf{All} \\
\toprule
\multirow{3}*{\textbf{LLaMa-7B}} & CoT-FSP  & 11.90 & 44.69 & 31.80 & 56.20 & 59.65 & 70.28 & 43.00 & 57.05  \\
    & CoT-FT      & 30.71 & 53.19 & 42.30 & 55.70 & 69.09 & 77.05 & 54.17 & 64.36  \\
    & CoA  & \textbf{35.71} & \textbf{56.36} & \textbf{51.10} & \textbf{67.59} & \textbf{80.51} & \textbf{85.94} & \textbf{68.33} & \textbf{75.98} \\
\midrule
\multirow{4}*{\textbf{LLaMa-2-7B}} & CoT-FSP  & 16.38 & 47.85 & 38.40 & 52.41 & 63.39 & 82.03 & 43.33 & 60.53 \\
    & CoT-FT      & 35.33 & 57.18 & 48.20 & 66.08 & 74.41 & 85.23 & 65.00 & 73.03  \\
    & Toolformer  & 17.59 & 48.55 & 37.10 & 47.34 & 58.46 & 79.54 & 50.67 & 59.81 \\
    & CoA  & \textbf{37.83} & \textbf{57.61} & \textbf{51.70} & \textbf{72.15} & \textbf{82.48} & \textbf{86.48} & 73.17 & \textbf{78.89} \\
\midrule
\multirow{8}*{\textbf{LLaMa-2-Chat-7B}} & CoT-FSP  & 24.03 & 54.14 & 51.30 & 71.90 & 72.44 & 85.41 & 74.00 & 76.32 \\
    & CoT-FT      & 35.41 & 59.00 & 46.90 & 58.23 & 72.24 & 85.41 & 73.00 & 73.37  \\
    & CoT-FT (no ASDiv)  & 36.19 & 44.93 & 35.30 & 38.48 & 52.95 & 61.21 & 77.67 & 59.61 \\
    & Toolformer  & 23.65 & 50.85 & 48.80 & 61.01 & 69.09 & 81.85 & 68.50 & 70.85  \\
    & Toolformer - Math & 36.01 & 59.18 & 47.60 & 58.99 & 72.44 & 85.94 & 75.50 & 74.43 \\
    & CoA & 38.29 & \textbf{59.57} & \textbf{54.20} & \textbf{72.41} & \textbf{81.89} & \textbf{88.26} & 83.00 & \textbf{82.13} \\
    & CoA (no ASDiv)  & \textbf{39.73} &  54.19  &  44.40  &  54.18  &  73.62  &  73.49  & \textbf{85.33} & 73.27 \\
    & CoA (no Tool) & 35.03 & 58.79 & 51.50 & 68.10 & 74.21 & 86.48 & 77.67 & 77.38 \\
\midrule
\multirow{5}*{\textbf{LLaMa-2-Chat-70B}} & CoT-FSP  & 56.18 & 65.94 & 70.60 & 86.08 & 89.17 & 92.88 & 84.50 & 88.23 \\
            & CoT-FT   & 60.50 & 70.24 & 70.40 & 81.52 & 87.60 & 92.35 & 89.17 & 88.18  \\
            & Toolformer & 52.54 & 69.07 & \textbf{73.60} & \textbf{86.84} & 89.76 & 91.46 & 81.50 & 87.26 \\
            & Toolformer - Math & 61.03 & 70.59 & 73.20 & 85.57 & 91.34 & 91.99 & 92.00 & 90.60 \\
            & CoA  & \textbf{62.32} & \textbf{71.89} & 73.40 & 86.33 & \textbf{94.49} & \textbf{93.06} & \textbf{92.33} & \textbf{91.91} \\
\midrule
\textbf{GPT-J} & Toolformer  & - & 40.4 & 29.4 & - & - & - & - & 44.0  \\
\bottomrule
\end{tabular}
}
\caption{Mathematical reasoning evaluation results.}
\label{tab:math_results_full}
\end{table*}

\begin{table*}[t]
\centering
\resizebox{0.9\textwidth}{!}{
\smallskip\begin{tabular}{llcccccc}
\toprule
\multirow{2}*{\textbf{Model}} & \multirow{2}*{\textbf{Method}}  & \multicolumn{3}{c}{\textbf{HotpotQA}} & \multirow{2}*{\textbf{WebQ.}} & \multirow{2}*{\textbf{NaturalQ.}} & \multirow{2}*{\textbf{TriviaQA}} \\
            \cmidrule(lr){3-5}
 & & \textbf{Bridge} & \textbf{Comparison} & \textbf{All} &  &  &   \\
\toprule
\multirow{4}*{\textbf{LLaMa-2-7B}} & CoT-FSP  & 14.43 & 45.26 & 20.62 & 33.96 & 33.35 & 56.95 \\
    & CoT-FT      & 14.85 & 57.36 & 23.39 & 31.50 & 26.93 & 52.32 \\
    & Toolformer  & 14.12 & 42.76 & 20.35 & \textbf{37.11} & 34.49 & 57.79 \\
    & CoA & \textbf{22.00} & \textbf{57.43} & \textbf{29.12} & 34.60 & \textbf{38.28} & \textbf{58.28} \\
\midrule
\multirow{6}*{\textbf{LLaMa-2-Chat-7B}} & CoT-FSP & 11.69 & 45.46 & 18.47 & 34.65 & 30.91 & 53.48 \\
    & CoT-FT      & 14.24 & 56.69 & 22.77 & 33.51 & 25.40 & 51.05 \\
    & Toolformer  & 12.99 & 44.59 & 20.00 & 36.22 & 30.22 & 54.15 \\
    & Toolformer - Wiki & 15.68 & 56.42 & 23.86 & \textbf{36.61} & 32.96 & 55.08 \\
    & FireAct     & 19.18 & 54.14 & 26.20 & 36.02 & 35.87 & 52.96 \\
    & CoA  & \textbf{21.00} & \textbf{56.96} & \textbf{28.22} & 35.97 & \textbf{38.67} & \textbf{57.90} \\
\midrule
\multirow{5}*{\textbf{LLaMa-2-Chat-70B}}  & CoT-FSP    & 21.39 & 56.62 & 28.47 & 34.89 & 37.42 & 63.61 \\
                            & CoT-FT      & 23.84 & 63.95 & 31.90 & 34.15 & 39.75 & 62.28 \\
                            & Toolformer  & 22.24 & 56.09 & 29.04 & 37.16 & 40.42 & 64.31 \\
                            & Toolformer - Wiki & 26.38 & 63.82 & 33.90 & \textbf{37.70} & 41.25 & 66.64 \\
                            & CoA  & \textbf{27.61} & \textbf{64.09} & \textbf{34.94} & 36.37 & \textbf{43.57} & \textbf{69.08} \\
\midrule
\textbf{GPT-J} & Toolformer & - & - & - & 26.3 & 17.7 & 48.8 \\
\bottomrule
\end{tabular}
}
\caption{Wiki QA evaluation results.}
\label{tab:wiki_results_full}
\end{table*}

\section{Implementation Details}
\label{apdx:implementation}
\begin{figure*}[t]
\centering
\includegraphics[width=0.8\textwidth]{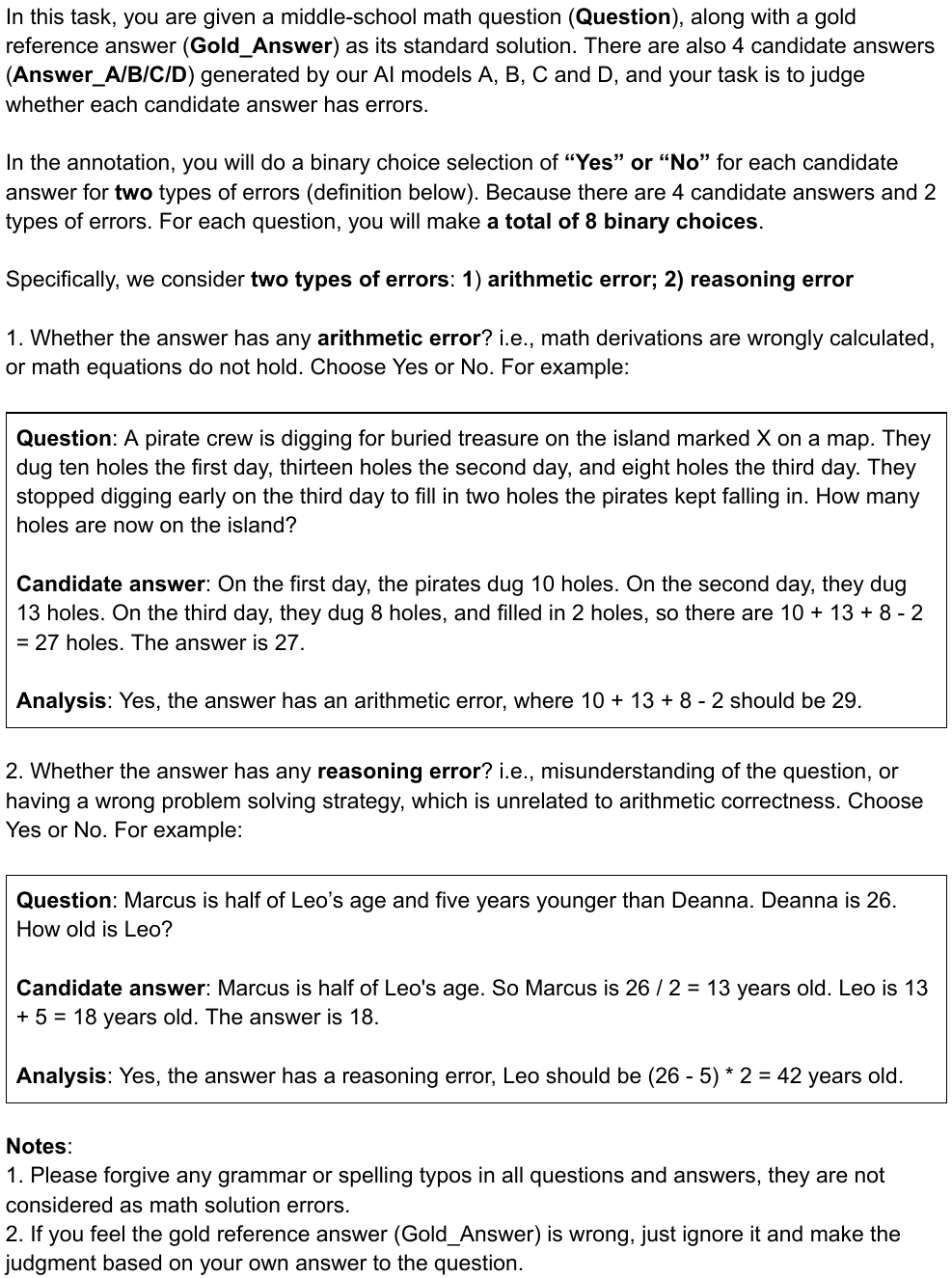}
\caption{Guideline for human evaluation on GSM8K mathematical reasoning.}
\label{guideline_human_eval}
\end{figure*}

\paragraph{Evaluation Details}
For mathematical reasoning evaluation, we extract the last number appeared in each model's answer, and check whether the number exactly match the gold reference. The accuracy is reported as the rate of such exact match across all QAs in a test set.
For Wiki QA evaluation, similar to mathematical reasoning, we extract the final answer of each model and calculate its exact match rate to the gold reference.
Specifically, the final answer is supposed to be the words after ``Action: finish['' for FireAct baseline, and words after ``The answer is '' for other models.
Our 8-shot in-domain examples used for the CoT-FSP baseline are shown in Table~\ref{tab:8_shot_math} and \ref{tab:8_shot_wiki}, which enables the model to provide answer with our required format for evaluation, \ie{}, stating its final answer after ``The answer is ''.
Our human evaluation on GSM8K is conducted by 5 internal domain experts from our research group.
For each math question, we provide the experts with the gold answer as reference, and ask them to evaluate each model answer in anonymous manner, \ie{}, experts do not know which model each answer comes from.
Two yes-or-no questions are asked for evaluating each model answer, including: a) whether the answer has any arithmetic error, and b) whether the answer has any reasoning error, and binary choices from the experts are collected to calculate the error rates of each model's generation.
We present our detailed instructions for human evaluation in Figure~\ref{guideline_human_eval}.
Our data collection protocol is approved by our organization in terms of ethics.

\paragraph{Model Training}
We fine-tune our models with batch size $8$ and learning rate $2e^{-5}$ and $1e^{-5}$ for 7B and 70B model sizes, respectively, using cosine learning rate scheduler with warm-up step $10$.
We use AdamW \citep{loshchilov2018decoupled} optimizer for all our fine-tuning experiments, with $\beta_{1}$, $\beta_{2}$ and $\epsilon$ set to $0.9$, $0.95$ and $1e^{-8}$, respectively.
Training weight decay is set to $0.1$.
For mathematical reasoning, we use a total of $400$ training steps, and get the best model checkpoints (with highest validation scores) at step $240$ and $200$ for 7B and 70B model sizes.
For Wiki QA domain, we adjust the total training steps to $500$, and get the best checkpoints at step $450$ and $300$ for 7B and 70B models.
Therefore, only $\sim$2K and $\sim$3K QAs are required in practice for fine-tuning our models in math and Wiki QA domains.
The training of our 7B and 70B models is based on 8 and 64 NVIDIA A100-SXM4 (80GB) GPUs, with training time about 2 and 5 hours per model, respectively.

\section{Full Experimental Results}
\label{apdx:full_results}
Table~\ref{tab:math_results_full} and \ref{tab:wiki_results_full} show the full results of our experiments on math and Wiki QA domains.
Our method of CoA achieves consistent improvements over baselines across various LLaMa model versions (LLaMa, LLaMa-2 and LLaMa-2-Chat), model sizes (7B and 70B), and domain benchmarks.
This shows great potential of our method being generalized to new model backbones and reasoning tasks.
We also present results on GSM8K subsets according to varying numbers of gold reasoning steps in Table~\ref{tab:math_steps}, where we confirm that CoA has more robust long chain-of-thought reasoning accuracy.

\paragraph{Fine-Tuning Data Balance}
In the mathematical reasoning domain, we also validate the importance of using fine-tuning data that is balanced across different reasoning steps.
Specifically, we conduct an ablation study on CoT-FT and CoA seeded with LLaMa-2-Chat-7B model, by removing the single-step QA samples of ASDiv from the fine-tuning data (\textbf{no ASDiv}).
We find that CoT-FT (no ASDiv) and CoA (no ASDiv) turn out to be biased towards multi-step reasoning, where they achieve better performance on GSM8K and MultiArith that contain mainly multi-step QAs, but suffer from severe performance degradation on other datasets that contain many single-step math problems.
This demonstrates that maintaining a good balance of single-step and multi-step reasoning data is important for adapting LLMs to be robust reasoners.


\paragraph{More Prompting Baselines}
We also compare our CoA reasoning method to more prompting-based methods PAL \citep{gao2023pal} and DECLARATIVE \citep{he2023solving}, which use few-shot coding demonstrations to prompt math solutions as Python or declarative programs. Table~\ref{tab:prompting_baseline} shows our comparison results on the GSM8K dataset, where all methods are seeded with LLaMa-2-Chat-7B. Without seeding with dedicated coding models (\eg{}, code-davinci-002), PAL and DECLARATIVE get far lower accuracy on GSM8K, which significantly under-perform our CoA method, and even ordinary CoT-FSP.

In contrast, our CoA method relies less on artificial demonstrations and distributional closeness of the seed LLM to target tasks, as CoA fine-tunes the LLM agent on pre-defined abstract reasoning chains, acquired from simple rewriting of natural language reasoning traces. Consequently, CoA is flexible in various generation formats, \eg{}, code and plain text, and generalizes well from mathematical reasoning to open-domain QA, which is a very different type of reasoning task. This indicates our method’s generalizability to novel reasoning schemes required by a new domain.

\begin{table}[t]
\centering
\resizebox{1.0\columnwidth}{!}{
\smallskip\begin{tabular}{lccccc}
\toprule
\multirow{2}*{\textbf{Method}}  &  \multicolumn{5}{c}{\textbf{Gold Reasoning Step}}   \\
                                        \cmidrule(lr){2-6}
                                & $\leq2$ & $3$ & $4$ & $5$ &  $>5$   \\
\toprule
CoT-FSP         & 42.9 & 26.3 & 18.0 & 10.9 & 3.6   \\
CoT-FT          & 55.5 & 42.6 & 25.8 & 19.0 & 10.8   \\
\midrule
\multirow{2}*{CoA}  & \textbf{55.8} & \textbf{44.4} & \textbf{32.5}  & \textbf{25.3} & \textbf{15.1} \\
& +0.3 & +1.8 & +6.7 & +6.3 & +4.3 \\
\bottomrule
\end{tabular}
}
\caption{Stratified LLaMa-2-Chat-7B evaluation results on GSM8K with different gold reasoning steps. The last row reports absolute accuracy improvement of our CoA method compared to CoT-FT baseline.}
\label{tab:math_steps}
\end{table}

\begin{table}[t]
\centering
\begin{tabular}{lc}
\toprule
\textbf{Method}  &  \textbf{Accuracy}   \\
\toprule
CoT-FSP     & 24.03 \\
PAL         & 20.55 \\
DECLARATIVE & 9.86  \\
CoA & \textbf{38.29} \\
\bottomrule
\end{tabular}
\caption{Comparison of CoA to prompting-based methods on GSM8K, seeded with LLaMa-2-Chat-7B.}
\label{tab:prompting_baseline}
\end{table}

\begin{table*}[t]
\centering
\resizebox{1.0\textwidth}{!}{
\smallskip\begin{tabular}{l}
\toprule
\textbf{Q}: There are 15 trees in the grove. Grove workers will plant trees in the grove today. After they are done, there will be 21 trees. How many trees will the grove \\
workers plant today? \\
\textbf{A}: There are 15 trees originally. Then there were 21 trees after some more were planted. So there must have been 21-15=6. The answer is 6. \\
\textbf{C}: There are 15 trees originally. Then there were 21 trees after some more were planted. So there must have been [21 - 15 = y1]. The answer is y1. \\
\\
\textbf{Q}: The flowers cost \$9, the clay pot costs \$20 more than the flower, and the bag of soil costs \$2 less than the flower. How much does it cost to plant the flowers? \\
\textbf{A}: The clay pot costs \$20 + \$9 = \$29. The bag of soil costs \$9 - \$2 = \$7. The cost to plant the flowers is \$9 + \$29 + \$7 = \$45. The answer is 45. \\
\textbf{C}: The clay pot costs [20 + 9 = y1]. The bag of soil costs [9 - 2 = y2]. The cost to plant the flowers is [9 + y1 + y2 = y3]. The answer is y3. \\
\\
\textbf{Q}: From March to August, Sam made \$460 doing 23 hours of yard work. However, from September to February, Sam was only able to work for 8 hours. If Sam \\
is saving up to buy a video game console that costs \$600 and has already spent \$340 to fix his car, how many more hours does he need to work before he can buy \\
the video game console? \\
\textbf{A}: Sam makes \$460 / 23 hrs = \$20/hr. From September to February, Sam made 8hrs x \$20/hr = \$160. From March to February, Sam made a total of \$460 + \$160 \\
= \$620. After fixing his car, he was left with \$620 - \$340 = \$280. Sam needs another \$600 - \$280 = \$320. Sam needs to work another \$320 / \$20/hr = 16 hours. \\
The answer is 16. \\
\textbf{C}: Sam makes [460 / 23 = y1] dollars per hour. From September to February, Sam made [8 * y1 = y2] dollars. From March to February, Sam made a \\
total of [460 + y2 = y3] dollars. After fixing his car, he was left with [y3 - 340 = y4]. Sam needs another [600 - y4 = y5] dollars. Sam needs to work \\
another [y5 / y1 = y6] hours. The answer is y6. \\
\\
\textbf{Q}: There were nine computers in the server room. Five more computers were installed each day, from Monday to Thursday. How many computers are now in \\
the server room? \\
\textbf{A}: There were originally 9 computers. For each of 4 days, 5 more computers were added. So 5 * 4 = 20 computers were added. 9 + 20 is 29. The answer is 29. \\
\textbf{C}: There were originally 9 computers. For each of 4 days, 5 more computers were added. So [5 * 4 = y1] computers were added. [9 + y1 = y2]. \\
The answer is y2. \\
\\
\textbf{Q}: Of the 90 people on William's bus, 3/5 were Dutch. Of the 1/2 of the Dutch who were also American, 1/3 got window seats. What's the number of Dutch \\
Americans who sat at the windows? \\
\textbf{A}: On the bus, the number of Dutch people was 3/5 of the total number, a total of 3/5 x 90 = 54 people. Out of the 54 people who were Dutch, 1/2 were Dutch \\
Americans, a total of 1/2 x 54 = 27 people. If 1/3 of the passengers on the bus identifying as Dutch Americans sat at the windows, their number is 1/3 x 27 = 9. \\
The answer is 9. \\
\textbf{C}: On the bus, the number of Dutch people was 3/5 of the total number, a total of [3/5 * 90 = y1] people. Out of the Dutch people, 1/2 were Dutch \\
Americans, a total of [1/2 * y1 = y2] people. If 1/3 of the passengers on the bus identifying as Dutch Americans sat at the windows, their number \\
is [1/3 * y2 = y3]. The answer is y3. \\
\bottomrule
\end{tabular}
}
\caption{Prompting examples for fine-tuning data construction in mathematical reasoning domain. Given a question (Q) and a gold answer (A), LLaMa-70B is prompted to generate the re-writing of answer as abstract reasoning chain (C). Based on that, our method trains a LLM to generate the abstract chain based on the question, and the final answer is derived by reify the chain of reasoning with the domain tool (\ie{}, equation solver).}
\label{tab:rewrite_math}
\end{table*}

\section{Fine-Tuning Data Re-writing Details}
\label{apdx:rewrite}
Table~\ref{tab:rewrite_math} and \ref{tab:rewrite_wiki} show the prompting examples for fine-tuning data construction of our method.
We prompt LLaMa-70B to re-write existing math and Wiki QAs as abstract reasoning chains, which gets rid of data distillation from close-sourced LLMs, yet obtains data resources that enable more effective learning of multi-step reasoning.


\begin{table*}[t]
\centering
\resizebox{1.0\textwidth}{!}{
\smallskip\begin{tabular}{l}
\toprule
\textbf{Q}: Fritz von Brodowski was killed during what global war that lasted from 1939 to 1945? \\
\textbf{A}: The answer is World War II. \\
\textbf{W}: Fritz von Brodowski > Friedrich Wilhelm Konrad von Brodowski was controversially killed while in French custody during World War II. \\
\textbf{C}: Find the [war in which Fritz von Brodowski was killed -Wiki-> y1]. \\
\\
\textbf{Q}: Which tennis player won more Grand Slam titles, Henri Leconte or Jonathan Stark? \\
\textbf{A}: The answer is Jonathan Stark. \\
\textbf{W}: Henri Leconte > He won the French Open men's doubles title in 1984. Jonathan Stark (tennis) > During his career he won two Grand Slam doubles titles. \\
\textbf{C}: First identify the [number of Grand Slam titles Henri Leconte won -Wiki-> y1]. Then find out the [number of Grand Slam titles Jonathan Stark won -Wiki-> y2]. \\
\\
\textbf{Q}: The director of the romantic comedy ``Big Stone Gap'' is based in what New York city? \\
\textbf{A}: The answer is Greenwich Village. \\
\textbf{W}: Big Stone Gap (film) > Big Stone Gap is a 2014 American romantic comedy film directed by Adriana Trigiani. Adriana Trigiani > Adriana Trigiani is an \\
Italian American film director based in Greenwich Village. \\
\textbf{C}: First search the [director of romantic comedy ``Big Stone Gap'' -Wiki-> y1]. The name of this film's director is [y1 -NER(person)-> y2]. Then determine [y2 in \\ 
what New York city -Wiki-> y3]. \\
\\
\textbf{Q}: Are Randal Kleiser and Kyle Schickner of the same nationality? \\
\textbf{A}: The answer is yes. \\
\textbf{W}: Randal Kleiser > John Randal Kleiser (born July 20, 1946) is an American film director and producer. Kyle Schickner > Kyle Schickner is an American film \\
producer, writer, director, actor. \\
\textbf{C}: First find out the [nationality of Randal Kleiser -Wiki-> y1]. Then figure out the [nationality of Kyle Schickner -Wiki-> y2]. \\
\\
\textbf{Q}: Extras was created, written, and directed by Ricky Dene Gervais, an English comedian, actor, writer, producer, director, singer, and musician, born on which date? \\
\textbf{A}: The answer is 25 June 1961. \\
\textbf{W}: Ricky Gervais > Ricky Dene Gervais (born 25 June 1961) is an English comedian, actor, writer, producer, director, singer, and musician. \\
\textbf{C}: Search [when Ricky Dene Gervais was born -Wiki-> y1]. \\
\\
\textbf{Q}: Sameera Perera is a cricketer from what island country located southeast of the Republic of India and northeast of the Maldives? \\
\textbf{A}: The answer is Sri Lanka. \\
\textbf{W}: Sameera Perera > Sameera Perera (born 20 August 1988) is a Sri Lankan cricketer. \\
\textbf{C}: Identify the [country that cricketer Sameera Perera is from -Wiki-> y1]. \\
\\
\textbf{Q}: What screenwriter with credits for ``Evolution'' co-wrote a film starring Nicolas Cage and Téa Leoni? \\
\textbf{A}: The answer is David Weissman. \\
\textbf{W}: The Family Man > The Family Man is a 2000 American romantic comedy-drama film starring Nicolas Cage and Téa Leoni. David Weissman > His film credits \\
include ``The Family Man'' (2000), ``Evolution'' (2001), and ``When in Rome'' (2010). \\
\textbf{C}: First figure out the [film of Nicolas Cage and Téa Leoni -Wiki-> y1]. The name of this film is [y1 -NER(culture)-> y2]. Then find out [who wrote y2 with \\
credits for ``Evolution'' -Wiki-> y3]. \\
\\
\textbf{Q}: Ralph Hefferline was a psychology professor at a university that is located in what city? \\
\textbf{A}: The answer is New York City. \\
\textbf{W}: Ralph Hefferline > Ralph Franklin Hefferline was a psychology professor at Columbia University. Columbia University > Columbia University is a private Ivy \\
League research university in Upper Manhattan, New York City. \\
\textbf{C}: First identify the [university of psychology professor Ralph Hefferline -Wiki-> y1]. The university of this professor is [y1 -NER(group)-> y2]. Then figure \\
out [y2 is in what city -Wiki-> y3]. \\
\bottomrule
\end{tabular}
}
\caption{Prompting examples for fine-tuning data construction in Wiki QA domain. Given a question (Q), a gold answer (A) and its supporting Wikipedia articles (W), LLaMa-70B is prompted to generate an abstract reasoning chain (C) with Wikipedia searching and NER queries. Based on that, our method first trains a LLM to generate the abstract chain of queries based on the question, and then execute the queries by domain tools (\ie{}, Wikipedia search engine and NER toolkit). Finally, a second LLM is trained to generate the final answer based on the Wikipedia searching results (excluding intermediate NER results) in the reified chain of reasoning.}
\label{tab:rewrite_wiki}
\end{table*}

\begin{table*}[t]
\centering
\resizebox{1.0\textwidth}{!}{
\smallskip\begin{tabular}{l}
\toprule
\textbf{Q}: There are 15 trees in the grove. Grove workers will plant trees in the grove today. After they are done, there will be 21 trees. How many trees will the grove \\
workers plant today? \\
\textbf{A}: There are 15 trees originally. Then there were 21 trees after some more were planted. So there must have been 21-15=6. The answer is 6. \\
\\
\textbf{Q}: If there are 3 cars in the parking lot and 2 more cars arrive, how many cars are in the parking lot? \\
\textbf{A}: There are originally 3 cars. 2 more cars arrive. 3 + 2 = 5. The answer is 5. \\
\\
\textbf{Q}: The flowers cost \$9, the clay pot costs \$20 more than the flower, and the bag of soil costs \$2 less than the flower. How much does it cost to plant the flowers? \\
\textbf{A}: The clay pot costs \$20 + \$9 = \$29. The bag of soil costs \$9 - \$2 = \$7. The cost to plant the flowers is \$9 + \$29 + \$7 = \$45. The answer is 45. \\
\\
\textbf{Q}: Maddie wants to see how much her mom spends on coffee each week. She makes herself 2 cups of coffee per day. Each cup has 1.5 ounces of coffee beans. \\
A bag of coffee costs \$8 and contains 10.5 ounces of beans. How much does she spend on her coffee per week? \\
\textbf{A}: She uses 3 ounces of beans per day because 2 x 1.5 = 3. She uses 21 ounces of beans per week because 7 x 3 = 21. She buys 2 bags of beans per week \\
because 21 / 10.5 = 2. She spends \$16 on the beans per week because 2 x 8 = 16. The answer is 16. \\
\\
\textbf{Q}: There were nine computers in the server room. Five more computers were installed each day, from Monday to Thursday. How many computers are now in \\
the server room? \\
\textbf{A}: There were originally 9 computers. For each of 4 days, 5 more computers were added. So 5 * 4 = 20 computers were added. 9 + 20 is 29. The answer is 29. \\
\\
\textbf{Q}: From March to August, Sam made \$460 doing 23 hours of yard work. However, from September to February, Sam was only able to work for 8 hours. If Sam \\
is saving up to buy a video game console that costs \$600 and has already spent \$340 to fix his car, how many more hours does he need to work before he can buy \\
the video game console? \\
\textbf{A}: Sam makes \$460 / 23 hrs = \$20/hr. From September to February, Sam made 8hrs x \$20/hr = \$160. From March to February, Sam made a total of \$460 + \$160 \\
= \$620. After fixing his car, he was left with \$620 - \$340 = \$280. Sam needs another \$600 - \$280 = \$320. Sam needs to work another \$320 / \$20/hr = 16 hours. \\
The answer is 16. \\
\\
\textbf{Q}: Of the 90 people on William's bus, 3/5 were Dutch. Of the 1/2 of the Dutch who were also American, 1/3 got window seats. What's the number of Dutch \\
Americans who sat at the windows? \\
\textbf{A}: On the bus, the number of Dutch people was 3/5 of the total number, a total of 3/5 x 90 = 54 people. Out of the 54 people who were Dutch, 1/2 were Dutch \\
Americans, a total of 1/2 x 54 = 27 people. If 1/3 of the passengers on the bus identifying as Dutch Americans sat at the windows, their number is 1/3 x 27 = 9. \\
The answer is 9. \\
\\
\textbf{Q}: Leah had 32 chocolates and her sister had 42. If they ate 35, how many pieces do they have left in total? \\
\textbf{A}: Originally, Leah had 32 chocolates. Her sister had 42. So in total they had 32 + 42 = 74. After eating 35, they had 74-35=39. The answer is 39. \\
\bottomrule
\end{tabular}
}
\caption{Few-shot examples used for CoT-FSP baseline model in mathematical reasoning domain.}
\label{tab:8_shot_math}
\end{table*}

\begin{table*}[t]
\centering
\resizebox{1.0\textwidth}{!}{
\smallskip\begin{tabular}{l}
\toprule
\textbf{Q}: Fritz von Brodowski was killed during what global war that lasted from 1939 to 1945? \\
\textbf{A}: Fritz von Brodowski > Friedrich Wilhelm Konrad von Brodowski was controversially killed while in French custody during World War II. The answer is World War II. \\
\\
\textbf{Q}: Which tennis player won more Grand Slam titles, Henri Leconte or Jonathan Stark? \\
\textbf{A}: Henri Leconte > He won the French Open men's doubles title in 1984. Jonathan Stark (tennis) > During his career he won two Grand Slam doubles titles. \\
The answer is Jonathan Stark. \\
\\
\textbf{Q}: The director of the romantic comedy ``Big Stone Gap'' is based in what New York city? \\
\textbf{A}: Big Stone Gap (film) > Big Stone Gap is a 2014 American romantic comedy film directed by Adriana Trigiani. Adriana Trigiani > Adriana Trigiani is an \\
Italian American film director based in Greenwich Village. The answer is Greenwich Village. \\
\\
\textbf{Q}: Are Randal Kleiser and Kyle Schickner of the same nationality? \\
\textbf{A}: Randal Kleiser > John Randal Kleiser (born July 20, 1946) is an American film director and producer. Kyle Schickner > Kyle Schickner is an American film \\
producer, writer, director, actor. The answer is yes. \\
\\
\textbf{Q}: Extras was created, written, and directed by Ricky Dene Gervais, an English comedian, actor, writer, producer, director, singer, and musician, born on which date? \\
\textbf{A}: Ricky Gervais > Ricky Dene Gervais (born 25 June 1961) is an English comedian, actor, writer, producer, director, singer, and musician. The answer is 25 June 1961. \\
\\
\textbf{Q}: Sameera Perera is a cricketer from what island country located southeast of the Republic of India and northeast of the Maldives? \\
\textbf{A}: Sameera Perera > Sameera Perera (born 20 August 1988) is a Sri Lankan cricketer. The answer is Sri Lanka. \\
\\
\textbf{Q}: What screenwriter with credits for ``Evolution'' co-wrote a film starring Nicolas Cage and Téa Leoni? \\
\textbf{A}: The Family Man > The Family Man is a 2000 American romantic comedy-drama film starring Nicolas Cage and Téa Leoni. David Weissman > His film credits \\
include ``The Family Man'' (2000), ``Evolution'' (2001), and ``When in Rome'' (2010). The answer is David Weissman. \\
\\
\textbf{Q}: Ralph Hefferline was a psychology professor at a university that is located in what city? \\
\textbf{A}: Ralph Hefferline > Ralph Franklin Hefferline was a psychology professor at Columbia University. Columbia University > Columbia University is a private Ivy \\
League research university in Upper Manhattan, New York City. The answer is New York City. \\
\bottomrule
\end{tabular}
}
\caption{Few-shot examples used for CoT-FSP baseline model in Wiki QA domain.}
\label{tab:8_shot_wiki}
\end{table*}

